# A Robust Machine Learning Approach for Path Loss Prediction in 5G Networks with Nested Cross Validation

İbrahim Yazıcı[1], and Emre Gures[2]

[1]Türk Telekom R&D Department, Acıbadem, Istanbul, Turkey
[2]Faculty of Electrical Engineering Czech Technical University in Prague, Prague, Czech Republic
Email: [1]ibrahim.yazici@turktelekom.com.tr, guresemr@cvut.cz

*Abstract*— The design and deployment of fifth-generation (5G) wireless networks pose significant challenges due to the increasing number of wireless devices. Path loss has a landmark importance in network performance optimization, and accurate prediction of the path loss, which characterizes the attenuation of signal power during transmission, is critical for effective network planning, coverage estimation, and optimization. In this sense, we utilize machine learning (ML) methods, which overcome conventional path loss prediction models drawbacks, for path loss prediction in a 5G network system to facilitate more accurate network planning, resource optimization, and performance improvement in wireless communication systems. To this end, we utilize a novel approach, nested cross validation scheme, with ML to prevent overfitting, thereby getting better generalization error and stable results for ML deployment. First, we acquire a publicly available dataset obtained through a comprehensive measurement campaign conducted in an urban macro-cell scenario located in Beijing, China. The dataset includes crucial information such as longitude, latitude, elevation, altitude, clutter height, and distance, which are utilized as essential features to predict the path loss in the 5G network system. We deploy Support Vector Regression (SVR), CatBoost Regression (CBR), eXtreme Gradient Boosting Regression (XGBR), Artificial Neural Network (ANN), and Random Forest (RF) methods to predict the path loss, and compare the prediction results in terms of Mean Absolute Error (MAE) and Mean Square Error (MSE). As per obtained results, XGBR outperforms the rest of the methods. It outperforms CBR with a slight performance differences by 0.4 % and 1 % in terms of MAE and MSE metrics, respectively. On the other hand, it outperforms the rest of the methods with clear performance differences. Ultimately, the paper presents ML deployment with the novel approach from which network planning, resource optimization, and performance improvement in wireless communication systems will benefit as well according to the obtained results.

## I. INTRODUCTION

The fifth-generation (5G) of mobile networks has revolutionized wireless communication with its promise of high data rates, increased efficiency, and improved network performance [1]–[3]. To fully leverage the capabilities of 5G, accurate path loss prediction plays a vital role in tasks such as network planning, coverage estimation, and optimization.

The path loss prediction in wireless communication systems traditionally relies on two main types of models: deterministic models [4], [5] and empirical models [6], [7]. Deterministic path loss models are based on the physical principles of wave propagation and use computational techniques such as ray tracing or finite-difference time-domain (FDTD) simulations to predict the path loss. These models take into account detailed environmental information, including geometry, material properties, and electromagnetic characteristics, to predict the path loss. By accurately modelling the physical mechanisms of wave transmission, reflection, and diffraction, deterministic models offer the potential for high accuracy in path loss prediction. However, their drawbacks include the need for extensive computational resources and time, making them less practical for real-time applications. Additionally, deterministic models often require subjective parameter settings and assumptions, which may introduce uncertainties and challenges in capturing the full complexity of wireless propagation in different scenarios.

In contrast, empirical models are statistical models that derive their parameters from measured data collected in specific propagation environments. These models rely on the analysis of large-scale measurement data to establish statistical relationships between the path loss and various propagation parameters. These models are relatively simple and computationally efficient, making them attractive for quick estimations and initial network planning. However, their accuracy may be limited when applied to generalized wireless environments that differ significantly from the conditions in which the models were derived. Empirical models often assume simplified conditions, and may not capture the full complexity of wave propagation effects, such as scattering, diffraction, or variations in environmental characteristics. Both deterministic and empirical models have their strengths and weaknesses in the path loss prediction [8]. The deterministic models can provide high accuracy when detailed environmental information is available, but their computational complexity and reliance on assumptions may restrict their practicality. On the other hand, the empirical models offer simplicity and efficiency but they may lack accuracy in generalized scenarios. In recent years, machine learning (ML) has emerged as a promising alternative approach for path loss prediction. ML models have potential to overcome the limitations of traditional models by leveraging comprehensive datasets and powerful algorithms to learn patterns and relationships directly from data, leading to accurate and efficient path loss predictions in diverse wireless propagation environments.

By harnessing the capabilities of ML algorithms and leveraging large-scale datasets, ML-based models have the potential to surpass traditional methods in terms of accuracy and computational efficiency [9]–[11]. Unlike empirical models and deterministic methods, ML models can automatically learn complex patterns and relationships from data, enabling them to capture the intricacies of wireless propagation in a more precise and adaptable manner. Moreover, ML-based approaches can handle diverse environmental conditions, frequency bands, and propagation scenarios, making them suitable for the complex and dynamic nature of wireless networks. In the existing literature, several research studies have explored the application of ML methods for path loss prediction in different wireless communication environments.

Ref. [12] presents a study on predicting path loss models in urban cellular networks using ML techniques. The research evaluates the performance of three machine learning algorithms, Support Vector Regression (SVR), Random Forest (RF), and K-Nearest Neighbor (KNN), for path loss prediction in a Long Term Evolution (LTE) network scenario at 2.1 GHz. The evaluation encompasses both line-of-sight (LOS) and non-line-of-sight (NLOS) propagation conditions, utilizing a path loss dataset generated from ray tracing simulations. The study compares the performance of the ML methods with the widely used COST231 Walfisch-Ikegami empirical model. The findings highlight the better performance achieved by all ML algorithms in path loss prediction, with root-mean-square errors ranging from 2.1-2.2 dB for LOS and 3.4-4.1 dB for NLOS locations. Ref. [13] introduces application of various ML-based approaches for the path loss prediction in a smart campus environment. The study demonstrates better performance of ML algorithms, including artificial neural network (ANN) and RF, compared to the conventional COST-231 Hata model.

Ref. [14] presents a path loss prediction method for urban areas using convolutional neural networks (CNN) and data extracted from online sources, such as OpenStreetMap and other geographical information systems. The approach incorporates top-view images and building footprint data to improve the accuracy of path loss estimation. The proposed model is compared with ray tracing and exhibited faster prediction speed while maintaining accuracy in path loss prediction.

Ref. [15] studies the use of ML algorithms to predict path loss in tropical regions. The study includes a measurement campaign at broadcasting stations in Akure metropolis, Nigeria, and demonstrates the accuracy and feasibility of ML-based models. The study also proposes a data expansion framework that combines ML algorithms and classical models to improve prediction performance in scenarios with limited measured data at new frequencies. The findings of this study show the potential of ML algorithms for path loss prediction in tropical regions. However, the study also identifies the need for additional research to address data availability issues, improve accuracy, and enhance computational efficiency.

Ref [16] presents a study on the path loss prediction in aircraft cabin environments using different ML methods. The models are compared with measured data at frequencies of 2.4 GHz, 3.52 GHz, and 5.8 GHz. The results demonstrate that the ML-based approaches, including BPNN, SVR, RF, and AdaBoost, outperform the classical log-distance model in terms of prediction accuracy at these frequencies. The proposed data expansion method further enhances prediction accuracy, especially when there are limited measurement samples at the 5.8 GHz.

Ref. [17] proposes an ML-based path loss prediction model for urban environments at millimeter wave frequencies. The model utilizes 28 GHz measurements from Manhattan, incorporating street clutter features from LiDAR data and compressed building information. The suggested approach achieves improved prediction accuracy compared to statistical and 3GPP models, with a root-mean-square error (RMSE) of 4.8 ± 1.1 dB. The model employs linear ML algorithms, such as Elastic-net regression and SVR with radial basis function (RBF) kernel, to enhance performance in extrapolation and robustness against overfitting.

In nearly most of the studies that used ML methods for path loss prediction problem utilize conventional cross validation scheme in their ML deployments. In this scheme, ML method tunes its parameters and evaluates model performance with the same data by using the same validation dataset, and thereby leaking the learned information from the same data into the model. Consequently, the chosen parameters make the model biased towards the dataset used, resulting in overfitting where the model excessively adapts to the specific dataset. However, nested cross validation scheme differs from the conventional one in the manner that it does not use the same dataset for both model selection and parameter selection, thereby avoiding the information leakage from the dataset into the model. This study is, to our knowledge, the first to use this novel approach for ML deployment in the path loss prediction.

By advancing the field of the path loss prediction through the application of ML with a novel cross validation scheme, this research aims to facilitate more accurate network planning, resource optimization, and performance improvement in wireless communication systems.

This paper is motivated by the advantages of ML over traditional approaches for path loss prediction and the growing need for accurate path loss prediction in the context of 5G networks. It focuses on the development and evaluation of an ML-based path loss prediction model. In the rest of the paper, we briefly introduced path loss prediction through ML methods in Section II, and ML deployment stages in Section III. Accordingly, we discuss results obtained in Section IV, and finally make concluding remarks in Section V.

## II. PATH LOSS PREDICTION BY MACHINE LEARNING

The path loss prediction is one of the prominent areas for performance optimization of wireless network systems. In next generation networks where multiple requirements such as deployments of heterogeneous cells are needed, and reliability are increased, the path loss prediction gains more attraction to provide certain requirements in the network systems. Hence, the path loss prediction should better be handled with the aim of promising solutions. To this end, we propose ML method solutions for the path loss prediction by framing it into a supervised ML problem.

In general, three types of learning are available in ML which are supervised, unsupervised, and reinforcement learning. In this paper, supervised learning which outputs are predicted according to pre-defined input(s)/feature(s). Output type may be either binary/categorical-valued or continous-valued, and these output types determine the task of supervised learning. If the output type is binary/categorical, then the supervised learning task is classification. On the other hand, the supervised learning task is regression when the output type is continuous. We use regression in our ML deployment for predicting the path loss for a wireless network in this paper.

There are several prominent ML methods in the literature, and we utilize SVR, CBR, ANN, XGB, RF with nested cross validation scheme in order to provide better generalization ability for the ML methods for the prediction task. We utilize nested cross validation scheme so as to separate model evaluation and tuning the model parameters with the same data, thereby preventing leaking the learned information from the same data into the model. By this way, obtained results become less affected by overfitting, and the model will have more reliable generalization error than conventional cross validation scheme.

*A. Support Vector Regression (SVR)*

Support Vector Machine (SVM) is a statistical learning theory-based method that utilizes hinge loss and kernel tricks, and was originally proposed for classification task [18]. In SVM mechanism, gaps between the categories are tried to be maximally separated. Through kernel trick, feature space is mapped onto a higher dimensional space to perform classification task, and SVM problem is then framed into a constrained optimization problem. The problem finally is solved through quadratic optimization. With the same mechanism, SVM is used for regression tasks, and this is called SVR as it is used in this paper.

*B. CatBoost Regression (CBR)*

CBR method is a novel boosting method that has recently gained attraction for deployments. The method was proposed by Yandex research. Beside handling with continuous-valued features in a dataset, it handles categorical features effectively, and offers scalable and GPU computations with the use of large datasets [19].

*C. Artificial Neural Network (ANN)*

ANN is one of the powerful ML method that that mimics learning process of human brains. In ANN, neurons are successively connected to each other to convey the information learned from the previous neurons throughout the whole neural network. Learning is performed in these layers which some activation fucntions are used there to process the information/data. With recent advancements, there are many neural network types which Recurrent Neural Networks, Convolutional Neural Networks, Long-Short Term Memory are some landmark examples of them. We use a simple feed-forward ANN which comprises of one input, one hidden, and one output layer in its architecture. The ANN performs both regression and classification tasks, and it is determined by the type of the utilized activation function in the output layer. In this paper, we use the ANN for regression task.

*D. Extreme Gradient Boosting Regression (XGBR)*

Extreme gradient boosting method is one of the boosting methods that extends gradient boosting method, and scalable ensemble learning method [20]. The method can perform parallel computing on trees, it is insensitive to input scaling and adept in learning higher order interaction between features [21].

*E. Random Forest (RF)*

RF is one of bagging methods in ensemble learning strategy, and provides a single result by combining results of multiple decision trees. It makes use of random samples of a dataset with replacement to form decision tree sets, and they are trained independently from each other to reach a single result. RF creates uncorrelated forest of decisin tress by feature randomness and bagging, and performs regression or classification tasks.

III. MACHINE LEARNING DEPLOYMENT FOR THE PATH LOSS PREDICTION

In this section, we provide ML deployment stages for the path loss prediction problem. We will provide data collection, data processing, model training, performance comparison stages in brief.

*A. Data Collection*

The dataset used in this study was obtained through a measurement campaign conducted in an urban macro-cell scenario located in Beijing, China. The area selected for the campaign is a mixed-use area with a variety of buildings, including residential, commercial, and office buildings. The buildings are mostly lower than ten stories, with a few taller buildings scattered throughout. There are also several large pedestrian bridges that cross the roads in the area. The road signs are sparsely distributed, and there is an average tree density of approximately 6 meters along both sides of the roads.

Signals received from a TD-SCDMA base station operating at a frequency of 2021.4 MHz were considered during the measurement campaign. The TD-SCDMA Base Station had an antenna height of approximately 40 meters above the ground.

The data collection process described is a common method for collecting radio frequency (RF) signal data in urban areas. The omni-directional receive antenna is used to collect data from all directions, and the drive-test equipment is used to record the received signal power and the corresponding location information.

Post-processing of the recorded data enabled the calculation of path loss values, which were subsequently mapped to their respective locations. The resulting dataset encompasses crucial features, including longitude, latitude, elevation, altitude, cluster height, and distance, which were used as input features to predict path loss.

It is important to note that the dataset used in this study is publicly available and can be accessed by interested readers in Ref. [22].

*B. Data Processing*

In ML deployments, normalization is used for SVR and ANN methods since they are sensitive to feature scaling.

Normalization also affects convergence speed of the methods. We use zero-mean normalization for features given in Equation 1.

$$x'_i = \frac{x_i - \mu}{\sigma} \quad (1)$$

In Equation 1, normalized and original values of $i^{th}$ data point are represented by $x'_i$ and $x_i$, respectively. Accordingly, mean and standard deviation of relevant feature are represented by $\mu$ and $\sigma$, respectively.

*C. Model Training*

In this paper, we use nested cross validation scheme which prevents information leakege from model evaluation to parameter selection of this model, thereby avoiding overfitting. This scheme contributes to get better generalization error for ML methods. Nested cross validation utilizes several parts of train/test/validation splits within two loops. In the inner loop, each training set is utilized to fit model to maximize the score. Accordingly, validation set is used for directly maximizing the score. In the outer loop, generalization errors are estimated through test dataset splits. In the inner loop, the information leakage of model is prevented contrary to conventional cross validation scheme. Hence, the effect of overfitting issue is reduced.

Conventional and nested cross validation schemes are shown in Figures 1 and 2, respectively. In our deployment for nested cross validation, we use outer cross validation split by 6-fold, and the inner one by 4-fold.

*D. Performance comparison*

We use single output prediction with multiple input/feature strategy. We use the following features to predict path loss: longitude, latitude, elevation, altitude, cluster height, and distance. For performance comparisons of results of the ML methods used, Mean Absolute Error (MAE) and Mean Square Error (MSE) are used as given in Equations 2-3:

$$MAE = \frac{1}{N} \sum_{i=1}^{N} |y_i - \hat{y}_i| \quad (2)$$

$$MSE = \frac{1}{N} \sum_{i=1}^{N} (y_i - \hat{y}_i)^2 \quad (3)$$

In Equations 2-3, $N$ denotes the size of dataset used, $y_i$ and $\hat{y}_i$ correspond to actual and predicted values of $i^{th}$ example, respectively.

## IV. RESULTS AND DISCUSSION

We present obtained results for path loss prediction, and discuss the results in this section. The obtained results are compared in terms of MAE and MSE metrics in Table 1, and the performance differences of the methods are also presented in the table.

The XGBR method outperforms the other methods in terms of both the MAE and MSE metrics. The XGBR method produces errors of 2.41 and 10.64 for the MAE and MSE metrics, respectively.

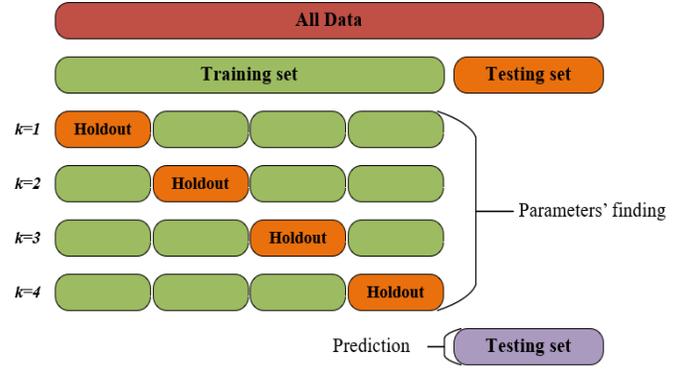

Fig. 1. Conventional cross validation scheme.

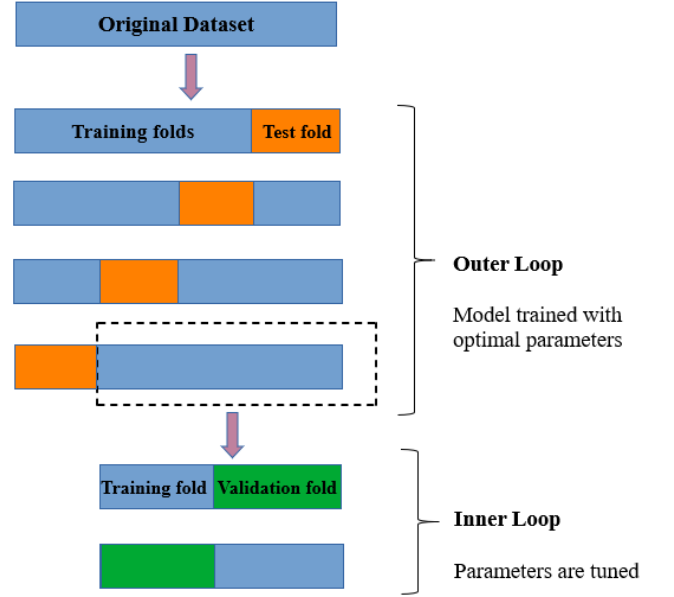

Fig. 2. Nested cross validation scheme.

TABLE 1. RESULTS

|  | MAE | MSE | Diff. (MAE) | Diff. (MSE) |
|---|---|---|---|---|
| SVR | 5.07 | 52.17 | 0.53 | 0.8 |
| CBR | 2.42 | 10.75 | 0.004 | 0.01 |
| ANN | 3.87 | 26.14 | 0.38 | 0.59 |
| XGBR | *2.41* | *10.64* | - | - |
| RFR | 2.97 | 15.23 | 0.19 | 0.3 |

It is followed by CBR method that produces errors by 2.42 and 10.75 in terms of MAE and MSE metrics, respectively. RF follows XGB and CBR methods, and it is followed by ANN. The SVR method performed the worst, with errors of 5.07 and 52.17 for the MAE and MSE metrics, respectively. Performance difference comparsions are also depicted in Figures 3-4.

The XGBR method outperformed the SVR method by 53% and 80% in terms of the MAE and MSE , respectively. The same differences with respect to ANN are 38% and 59%. The same differences with respect to RF are 19% and 30% , and this is 0.4 % and 1 % with respect to CBR.

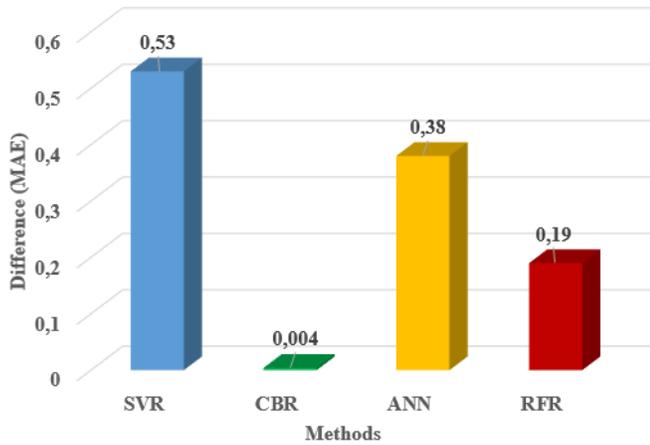

Fig. 3.  XGBR performance difference comparisons in MAE.

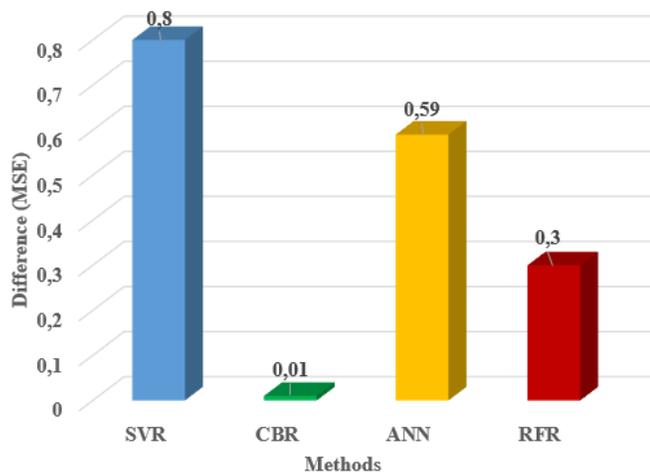

Fig. 4.  XGBR performance difference comparisons in MSE.

XGBR and CBR methods which two of them are boosting methods are the best perfroming methods with respect to the others. As per the results, while XGBR slighlty outperforms CBR method, the method superiorly outperforms the rest of the methods. It is noteworthy that two boosting methods outperform the other methods used. This shows the ability of boosting methods better handling of tabular type data since used dataset in this paper exhibits this characteristics. This is obvious from RF method's following just behind the best two methods. It is a bagging method that falls in ensemble learning type along with boosting methods. In general, ANN produces promising results, however, it is not the case in this study. This may be due to small dataset size since novel deep neural networks produce promising results, however, they are not suitable for the problem in this study.

## V. CONCLUSION

In this paper, we try to make the path loss prediction with ML deployment which , which overcome conventional path loss prediction models drawbacks. We deploy ML methods using a novel validation scheme, nested cross validation, to get better and more reliable results. Nested cross validation prevents infromation leakage from model evalaution to the model parameter tuning phase. Ultimately, this scheme prevents overfitting, thereby contributing to better generalization error for ML methods. Since the path loss prediction has landmark importance in optimizing wireless network performance, we approach to this problem with a novel scheme through deployment of ML methods which are SVR, CBR, XGB, ANN, and RF. In this study, we first acquire a publicly available dataset obtained through a comprehensive measurement campaign in an urban macro-cell scenario, enabling us to train and evaluate our novel ML scheme for path loss prediction in 5G wireless networks. Then, data processing for relevant ML methods is performed. This is followed by predictions. Finally, we compare obtained results in terms of MAE and MSE metrics. As per the obtained results, XGBR outperforms the rest of the methods. It outperforms CBR with a slight performance differences by 0.4 % and 1 % in terms of MAE and MSE metrics, respectively. On the other hand, it outperforms the rest of the methods with obvious performance differences. These results obtained show the efficiency of boosting methods for the problem used in this paper. Ultimately, the deployment of ML in the paper will enhance network planning, resource optimization, and network performance of wireless networks in real world use-cases. The future study may be to extend the same deployment for different dataset for the path loss predictions with different feature sets. Another future research direction may be to examine the effects of data size on the path loss prediction since the results may guide radio frequency planners in practice.

## REFERENCES

[1]  E. Gures, I. Shayea, A. Alhammadi, M. Ergen, and H. Mohamad, "A comprehensive survey on mobility management in 5G heterogeneous networks: Architectures, challenges and solutions," *IEEE Access*, vol. 8, pp. 195883–195913, 2020.
[2]  E. Gures, I. Shayea, M. Sheikh, M. Ergen, and A. A. El-Saleh, "Adaptive cell selection algorithm for balancing cell loads in 5G heterogeneous networks," *Alexandria Eng. J.*, vol. 72, pp. 621–634, 2023.
[3]  E. Gures *et al.*, "Load balancing in 5G heterogeneous networks based on automatic weight function," *ICT Express*, 2023.
[4]  E. Montiel, A. S. Aguado, and F. X. Sillion, "A radiance model for predicting radio wave propagation in irregular dense urban areas," *IEEE Trans. Antennas Propag.*, vol. 51, no. 11, pp. 3097–3108, 2003.
[5]  T. K. Sarkar, Z. Ji, K. Kim, A. Medouri, and M. Salazar-Palma, "A survey of various propagation models for mobile communication," *IEEE Antennas Propag. Mag.*, vol. 45, no. 3, pp. 51–82, 2003.
[6]  M. Ayadi, A. Ben Zineb, and S. Tabbane, "A UHF path loss model using learning machine for heterogeneous networks," *IEEE Trans. Antennas Propag.*, vol. 65, no. 7, pp. 3675–3683, 2017.
[7]  N. Moraitis, P. Constantinou, F. Perez Fontan, and P. Valtr, "Propagation measurements and comparison with EM techniques for in-cabin wireless networks," *EURASIP J. Wirel. Commun. Netw.*, vol. 2009, pp. 1–13, 2009.
[8]  Y. Zhang, J. Wen, G. Yang, Z. He, and J. Wang, "Path loss prediction based on machine learning: Principle, method, and data expansion," *Appl. Sci.*, vol. 9, no. 9, p. 1908, 2019.
[9]  İ. Yazici, I. Shayea, and J. Din, "A survey of applications of artificial intelligence and machine learning in future mobile networks-enabled systems," *Eng. Sci. Technol. an Int. J.*, vol. 44, p. 101455, 2023.
[10] E. Gures, I. Yazici, I. Shayea, M. Sheikh, M. Ergen, and A. A. El-Saleh, "A comparative study of machine learning-based load balancing in high-speed," *Alexandria Eng. J.*, vol. 72, pp. 635–647, 2023.
[11] E. Gures, I. Shayea, M. Ergen, M. H. Azmi, and A. A. El-Saleh, "Machine Learning Based Load Balancing Algorithms in Future Heterogeneous Networks: A Survey," *IEEE Access*, 2022.
[12] N. Moraitis, L. Tsipi, and D. Vouyioukas, "Machine learning-based methods for path loss prediction in urban environment for LTE networks," in *2020 16th international conference on wireless and mobile computing, networking and communications (WiMob)*, 2020, pp. 1–6.

[13] H. Singh, S. Gupta, C. Dhawan, and A. Mishra, "Path loss prediction in smart campus environment: Machine learning-based approaches," in *2020 IEEE 91st Vehicular Technology Conference (VTC2020-Spring)*, 2020, pp. 1–5.

[14] I. F. M. Rafie, S. Y. Lim, and M. J. H. Chung, "Path Loss Prediction in Urban Areas: A Machine Learning Approach," *IEEE Antennas Wirel. Propag. Lett.*, 2022.

[15] O. J. Famoriji and T. Shongwe, "Path Loss Prediction in Tropical Regions using Machine Learning Techniques: A Case Study," *Electronics*, vol. 11, no. 17, p. 2711, 2022.

[16] J. Wen, Y. Zhang, G. Yang, Z. He, and W. Zhang, "Path loss prediction based on machine learning methods for aircraft cabin environments," *Ieee Access*, vol. 7, pp. 159251–159261, 2019.

[17] A. Gupta, J. Du, D. Chizhik, R. A. Valenzuela, and M. Sellathurai, "Machine learning-based urban canyon path loss prediction using 28 ghz manhattan measurements," *IEEE Trans. Antennas Propag.*, vol. 70, no. 6, pp. 4096–4111, 2022.

[18] V. Vapnik, *Statistical learning theory*, 1st ed. Wiley-Interscience, 1998.

[19] L. Prokhorenkova, G. Gusev, A. Vorobev, A. V. Dorogush, and A. Gulin, "CatBoost: unbiased boosting with categorical features," *Adv. Neural Inf. Process. Syst.*, vol. 31, 2018.

[20] T. Chen and C. Guestrin, "Xgboost: A scalable tree boosting system," in *Proceedings of the 22nd acm sigkdd international conference on knowledge discovery and data mining*, 2016, pp. 785–794.

[21] F. B. Mismar and B. L. Evans, "Partially blind handovers for mmWave new radio aided by sub-6 GHz LTE signaling," in *2018 IEEE International Conference on Communications Workshops (ICC Workshops)*, 2018, pp. 1–5.

[22] "Dataset." https://github.com/charchitd/Path-Loss-Prediction-Based-on-Machine-Learning-Principle-Method-and-Data-Expansion/blob/master/Dataset/dat.csv. Accession Date: 19.07.2023